\pdfoutput=1

\documentclass[11pt]{article}

\usepackage[]{acl}

\usepackage{times}
\usepackage{latexsym}
\usepackage{hyperref}

\usepackage[T1]{fontenc}

\usepackage[utf8]{inputenc}

\usepackage{microtype}

\usepackage{amsmath,amsfonts,bm}
\usepackage{booktabs} %
\usepackage{url}
\usepackage{graphicx}
\usepackage{array}
\usepackage{color}
\usepackage{makecell}
\usepackage{multirow}
\usepackage{xspace}
\usepackage{colortbl}
\usepackage{subcaption}

\usepackage{pifont}
\usepackage{graphics}
\newcommand{\xmark}{\ding{55}}%

\newcommand{\ours}{\textsc{Eider}\xspace}

\newcommand{\edit}[1]{\textcolor{black}{#1}}

\newcommand{\start}[1]{\vspace{.3mm}\noindent{{\bf #1}.}}

\newcommand{\downv}{\vspace{-.0cm}}
\newcommand{\upv}{\vspace{-.0cm}}

\definecolor{gred}{RGB}{255,102,102}
\definecolor{gblue}{RGB}{51,102,255}
\definecolor{gyellow}{RGB}{244,180,0}
\definecolor{ggreen}{RGB}{15,157,88}
\definecolor{ggrey}{RGB}{115,115,115}
\definecolor{na}{gray}{0.9}
\definecolor{LightYellow}{RGB}{255,255,191}
\definecolor{OrangeRed}{rgb}{1.0, 0.27, 0.0}
\definecolor{midnightgreen}{rgb}{0.0, 0.29, 0.33}
\definecolor{darkgreen}{rgb}{0.0, 0.42, 0.24}

\newcommand{\colorR}[1]{\textcolor{gred}{\textbf{#1}}}
\newcommand{\colorG}[1]{\textcolor{ggreen}{\textbf{#1}}}
\newcommand{\colorB}[1]{\textcolor{gblue}{\textbf{#1}}}

\usepackage{amsmath,amsfonts,bm}

\def\eqref#1{equation~\ref{#1}}

\def\1{\bm{1}}

\def\vb{{\bm{b}}}

\def\vz{{\bm{z}}}

\def\mA{{\bm{A}}}

\def\mH{{\bm{H}}}

\def\mW{{\bm{W}}}

\DeclareMathAlphabet{\mathsfit}{\encodingdefault}{\sfdefault}{m}{sl}
\SetMathAlphabet{\mathsfit}{bold}{\encodingdefault}{\sfdefault}{bx}{n}

\def\sR{{\mathbb{R}}}

\title{\ours: Empowering Document-level Relation Extraction with Efficient Evidence Extraction and Inference-stage Fusion}

\author{Yiqing Xie, Jiaming Shen, Sha Li, Yuning Mao, Jiawei Han \\
University of Illinois at Urbana-Champaign, IL, USA \\
\{xyiqing2, js2, shal2, yuningm2, hanj\}@illinois.edu 
}

\begin{document}
\maketitle
\begin{abstract}

Document-level relation extraction (DocRE) aims to extract semantic relations among entity pairs in a document. 
Typical DocRE methods blindly take the full document as input, while a subset of the sentences in the document, noted as the evidence, are often sufficient for humans to predict the relation of an entity pair.
In this paper, we propose an evidence-enhanced framework, \ours, that empowers DocRE by efficiently extracting evidence and effectively fusing the extracted evidence in inference.\footnote{Our code is available at \url{https://github.com/Veronicium/Eider}}
\edit{We first jointly train an RE model with a lightweight evidence extraction model, which is efficient in both memory and runtime.
Empirically, even training the evidence model on silver labels constructed by our heuristic rules can lead to better RE performance.}
We further design a simple yet effective inference process that makes RE predictions on both extracted evidence and the full document, then fuses the predictions through a blending layer. 
This allows \ours to focus on important sentences while still having access to the complete information in the document.
Extensive experiments show that \ours outperforms state-of-the-art methods on three benchmark datasets (e.g., by 1.37/1.26 Ign F1/F1 on DocRED). 
\end{abstract}

\section{Introduction}
Relation extraction~(RE) is the task of extracting semantic relations among entities within a given text, which has abundant applications such as knowledge graph construction, question answering, and biomedical text analysis \citep{yu-etal-2017-improved,Shi2019DiscoveringHI,trisedya-etal-2019-neural}. 
Prior studies mostly focus on predicting the relation between two entity mentions in a single sentence.
However, in reality, an entity may have multiple mentions throughout a document.
It is also common that a relation can only be inferred given multiple sentences as the context.
As a result, recent studies have been moving towards the more realistic setting of document-level relation extraction (DocRE) \citep{peng-etal-17-cross,DocRED-paper,GAIN}.

\begin{figure}[t]
    \centering
    \includegraphics[width=\linewidth]{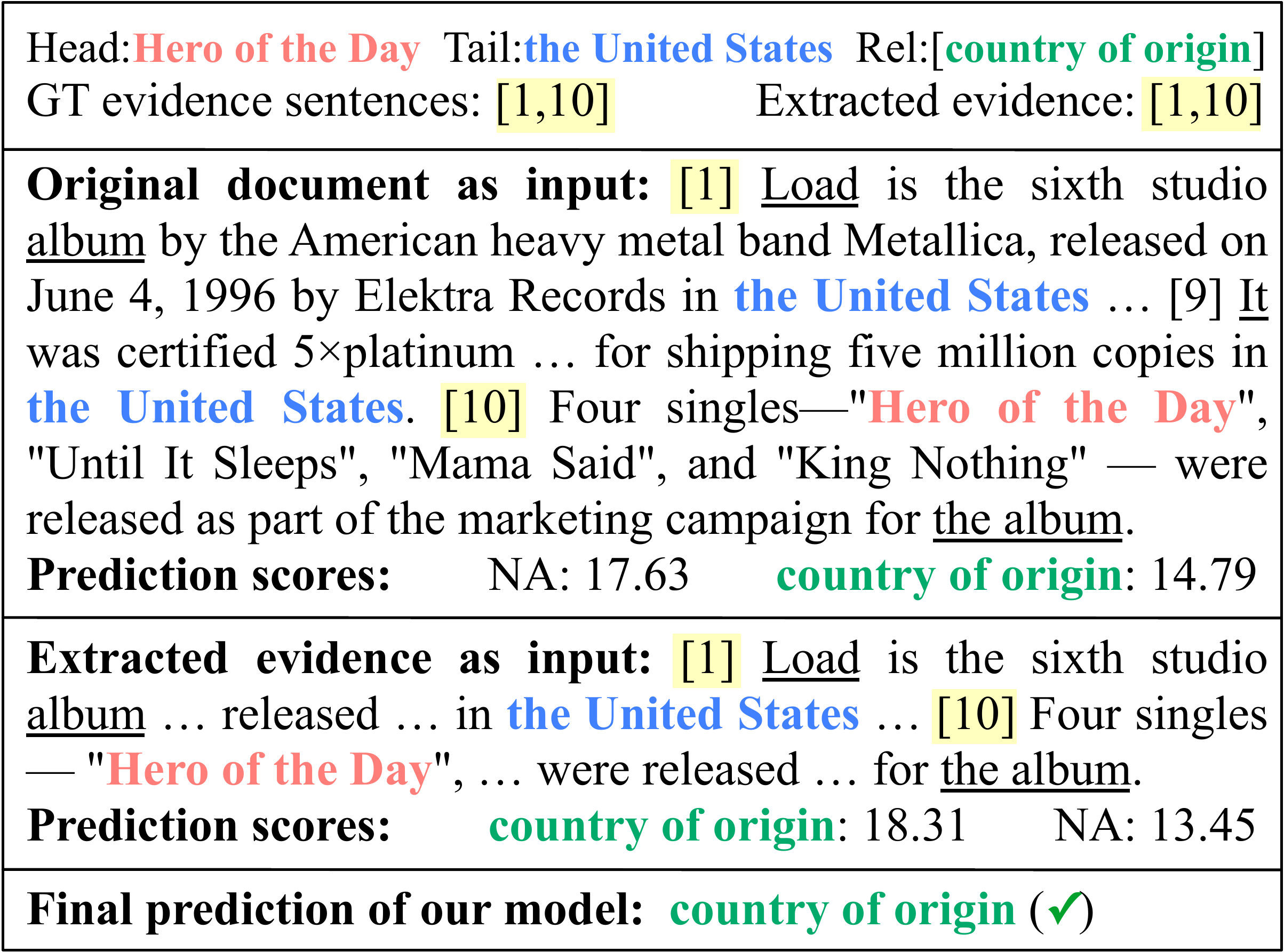}
    \upv
    \vspace{-0.3cm}
    \caption{ A test sample in the DocRED dataset \cite{DocRED-paper}, where the $i^{th}$ sentence in the document is marked with [i] at the start. Our model correctly predicts [1,10] as evidence, and if we only use the extracted evidence as input, the model can predict the relation ``country of origin'' correctly. }
    \label{fig_intro}
    \downv
    \vspace{-0.2cm}
\end{figure}

Unlike typical DocRE models that blindly take the whole document as input, a human may only need a few sentences to infer the relation of an entity pair.
For each entity pair, we define the minimal set of sentences required by human annotators to infer their relation as their \textit{evidence sentences}.
As shown in Figure~\ref{fig_intro}, to predict the relation between \textit{``Hero of the Day''} and \textit{``the United States''}, it is sufficient to know that \textit{Load (the album)} was released in \textit{the United States} from the $1^{st}$ sentence, and \textit{``Hero of the Day''} is a single of \textit{Load} from the $10^{th}$ sentence. In other words, the $1^{st}$ and $10^{th}$ sentences serve as the evidence to infer this relation.
Although the $9^{th}$ sentence also mentions \textit{``the United States''}, it is irrelevant to this specific relation. Including such irrelevant sentences in input might sometimes introduce noise to the model and be more detrimental than beneficial.

\edit{Despite the usefulness of evidence, few prior studies leverage it in a proper way \cite{E2GRE,3Sent}.  
In particular, \citet{E2GRE} extracts the evidence sentences together with RE but does not utilize them after extraction. Besides, it requires human-annotated evidence for training, and also suffers from massive memory usage and training time. 
Another work \cite{3Sent} trains an RE model solely on evidence sentences, which misses important information in the original document and fails to show improvements when paired up with pre-trained language models.}

In this paper, we propose an \textbf{e}v\textbf{id}ence-\textbf{e}nhanced Doc\textbf{R}E framework \ours, which \edit{efficiently} extracts evidence and effectively leverages the extracted evidence to improve DocRE.
During training, we enhance DocRE by jointly extracting relations and evidence using multi-task learning, \edit{which allows the two tasks to benefit from providing additional training signals for each other. There are two major challenges regarding evidence extraction.}
\edit{The first challenge is the memory and runtime overhead due to training an additional task. For example, a prior multi-task method \cite{E2GRE} needs over 14h and three consumer GPUs to train, while the individual RE model only takes around 90min on one GPU.
In comparison, \ours uses a simpler evidence extraction model, which can fit into a single GPU and only requires 95min runtime.}
The second challenge is that human-annotated evidence sentences are costly and heavily relying on them limits model applicability.
Therefore, we design several heuristic rules to construct silver labels in case the evidence annotation is unavailable.
We observe that \ours still improves RE performance when trained with our silver labels, and sometimes even performs on par with using gold labels.

With the evidence extracted, either by our rules or evidence extraction model, 
we propose to further enhance DocRE by utilizing the evidence in inference.
In the extreme case, if there is only one sentence related to the relation, one can make predictions solely based on this sentence and reduce the problem to sentence-level relation extraction.
One naive approach is thus to directly replace the original document with the extracted evidence \cite{3Sent}.
However, since no systems can extract evidence perfectly, solely relying on extracted sentences may miss important information and harm model performance in certain cases (see Table~\ref{tab:ablation}).
To avoid information loss, we fuse the prediction results of the original document and extracted evidence through a blending layer \cite{Blending}.
In this way, \ours pays more attention to the extracted important sentences, while still having access to all the information in the document. 
Empirical analysis demonstrates that removing either source would lead to degenerate performance.

We conduct extensive experiments on three widely-adopted DocRE benchmarks: DocRED \cite{DocRED-paper}, CDR \cite{CDR} and GDA \cite{GDA}.
Experiment results show that \ours achieves state-of-the-art performance on all the datasets.
Performance analysis further shows that the improvement of \ours is most significant on inter-sentence entity pairs, suggesting that leveraging evidence is especially effective in reasoning over multiple sentences.
In particular, \ours significantly improves the performance on entity pairs that require co-reference/multi-hop reasoning by 1.98/2.08 F1 on DocRED, respectively.

\start{Contributions}
(1) We propose an efficient joint relation and evidence extraction model that allows the two tasks to mutually enhance each other without heavily relying on evidence annotation.
(2) We design a simple and effective DocRE inference process enhanced by the extracted evidence, enabling more focus on the important sentences with no information loss.
(3) We demonstrate that our evidence-enhanced framework outperforms state-of-the-art methods on three DocRE datasets.
\begin{figure*}[ht]
    \centering
    \includegraphics[width=0.98\linewidth]{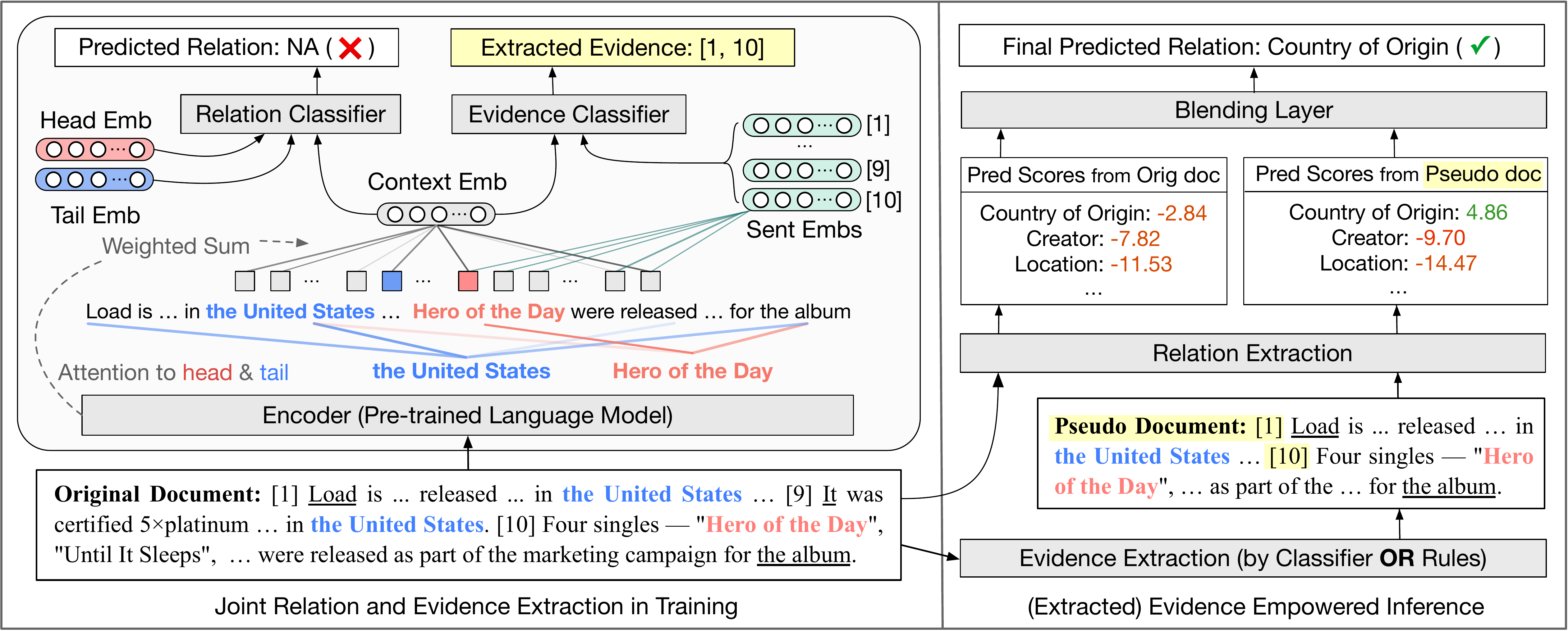}
    \upv
    \caption{The overall architecture of \ours.
    The left part illustrates the training stage and the right shows the inference stages of \ours. 
    We highlight \colorR{head entities}, \colorB{tail entities} and \colorbox{LightYellow}{extracted evidences.}}
    \label{fig:framework}
    \downv
\end{figure*}

\section{Problem Formulation}
Given a document $d$ comprised of $N$ sentences $\{s_n\}_{n=1}^N$, $L$ tokens $\{h_l\}_{l=1}^L$, $E$ named entities $\{e_i\}_{i=1}^E$ and all the proper-noun mentions of each entity, $\{m^i_j\}$, 
the task of document-level relation extraction (DocRE) is to predict the set of all possible relations between all entity pairs $(e_h, e_t)$ from a pre-defined relation set $\mathcal{R} \bigcup \{\text{NA}\}$. 
We refer to $e_h$ and $e_t$ as the head entity and tail entity, respectively.
A relation $r$ belongs to the positive class $\mathcal{P}^T_{h,t}$ if it exists between $(e_h, e_t)$ and otherwise the negative class $\mathcal{N}^T_{h,t}$.
For each entity pair $(e_h, e_t)$ that possesses a non-$\text{NA}$ relation, we define its \textit{evidence}\footnote{\small We use ``\textit{evidence sentence}'' and ``\textit{evidence}'' interchangeably throughout the paper.} $V_{h,t} = \{s_{v_k}\}_{k=1}^{K}$ as the subset of sentences in the document that are sufficient for human annotators to infer the relation.
Human annotation of evidence may or may not be given in training, depending on the datasets, but is not available in inference.

\section{Methodology}

An illustration of the framework of \ours is shown in Figure~\ref{fig:framework}.
In training, we jointly extract relation and evidence using multi-task learning, where the two tasks have their own classifier and share the base encoder (Sec.~\ref{sec:joint}).
In inference, we fuse the predictions on the original document and the extracted evidence using a blending layer (Sec.~\ref{sec:inference}).
In case the evidence annotation is not available, we also provide several heuristic rules to construct silver evidence labels as an alternative (Sec.~\ref{sec:rule}).

\subsection{Joint Relation and Evidence Extraction}
In our framework, we jointly train the relation extraction model with an evidence extraction model using multi-task learning.
As shown in Figure \ref{fig:framework}, the two tasks have their own classifier but share the base encoder.
Intuitively, tokens relevant to predicting the relation are essential in both models. By sharing the base encoder, the two tasks can provide additional training signals for each other and hence mutually enhance each other~\cite{Ruder2017AnOO}.

\smallskip
\start{Base Encoder}
\label{sec:joint}
We leverage pre-trained language models \cite{BERT} to encode the semantic meanings of each token in the document.
Specifically, given a document $d = [h_l]_{l=1}^{L}$, we insert a special token ``*'' before and after each entity mention $\{m^i_j\}$ and leverage the encoder to obtain the $s$-dim token embeddings $\mH = [\mathbf{h}_1,  ..., \mathbf{h}_L], \mathbf{h}_l \in \sR^{s}$ and the cross token attention $\mA \in \mathbb{R}^{L \times L}$:
\begin{equation}
    \mH, \mA = \text{Encoder}(\left[h_1,  ..., h_L\right]), \label{eq:encoder}
\end{equation}
where $\mA$ is the average of the attention heads in the last transformer layer~\cite{transformer}.
For each mention of an entity $e_i$, we use the embedding of the start symbol ``*'' as its mention embedding $\mathbf{m^i_j}$. 
Then, we obtain the embedding of entity $e_i$ by adopting LogSumExp pooling~\citep{n-ary,ATLOP} over the embeddings of all its mentions: $\mathbf{e}_{i} = \log \sum_{j} \exp ( \mathbf{m^i_j} )$.

To predict the relation of different entity pairs, a model may need to focus on different parts of the context. 
To capture the context relevant to each entity pair $(e_h, e_t)$, we compute its context embedding $\mathbf{c}_{h,t}\in \sR^{s}$ based on the attention matrix $\mA$ from the pre-trained encoder \cite{ATLOP}:
\begin{equation}
 \begin{aligned}
     \mathbf{c}_{h,t} &= \mH^T \frac{\mA_h \circ \mA_t}{\mA_h^T \mA_t},
 \end{aligned}
\label{eq:lop}
\end{equation}
where $\circ$ is the Hadamard product and $\mA_h \in \mathbb{R}^L$ is $e_h$’s attention to all the tokens in the document,
obtained by averaging $e_h$’s mention-level attention.
Similarly for $\mA_t$. The intuition is that tokens with high attention towards both $e_h$ and $e_t$ are important to both entities. Hence, these tokens are likely to be essential to the relation and should contribute more to the context embedding.

\smallskip
\start{Relation Classifier}
To predict the relation between an entity pair ($e_h$, $e_t$), we first compute their context-aware representations ($\mathbf{z_h}$, $\mathbf{z_t}$) by combining their entity embeddings ($\mathbf{e}_h$, $\mathbf{e}_t$) with their context embedding $\mathbf{c}_{h,t}$ and then utilize a bilinear function to calculate the logit of how likely a relation $r\in\mathcal{R}$ exists between $e_h$ and $e_t$:

\begin{equation}
 \begin{aligned}
     \vz_h &=\tanh \left( \mW_h \mathbf{e}_h + \mW_{c_h} \textbf{c}_{h,t}\right), \\
     \vz_t &=\tanh \left( \mW_t \mathbf{e}_t + \mW_{c_t} \textbf{c}_{h,t}\right), \\
     \mathbf{y}_r  &=  \vz_h  \mW_r \vz_t  + \vb_r, \\
 \end{aligned}
\end{equation}
where $\mW_h, \mW_t, \mW_{c_h}, \mW_{c_t}, \mW_r$ and $\vb_r$ are learnable parameters. 
As the model may have different confidence for different entity pairs, we apply the adaptive-thresholding loss~\citep{ATLOP}, which learns a dummy relation class $\mathrm{TH}$ that serves as the dynamic threshold for each entity pair:
\begin{equation}
    \mathbf{y}_{\mathrm{TH}}  = \vz_h  \mW_{\mathrm{TH}} \vz_t  + \vb_r.
\end{equation}
During inference, for each tuple $(e_h, e_t, r), r \in \mathcal{R}$, we obtain the prediction score: $S_{h,t,r}^{(O)} = \mathbf{y}_r - \mathbf{y}_{TH}$. 
Finally, we define our training objective for relation extraction as follows:
\begin{gather}
\small
\begin{aligned}
    \mathcal{L}_{RE} &= -\sum_{h \neq t} \sum_{r\in \mathcal{P}^T_{h,t}} \log \left( \frac{\exp\left(\mathbf{y}_r\right)}{\sum_{r' \in \mathcal{P}^T_{h,t} \cup \{\mathrm{TH}\}} \exp \left(\mathbf{y}_{r'}\right)} \right) \\
    &- \log \left( \frac{\exp\left(\mathbf{y}_{\mathrm{TH}}\right)}{\sum_{r' \in \mathcal{N}^T_{h,t} \cup \{\mathrm{TH}\}} \exp \left(\mathbf{y}_{r'}\right)} \right).
\end{aligned}
\raisetag{17pt}
\end{gather}

\start{Evidence Classifier}
In addition to the relation, we also predict whether each sentence $s_n$ is an evidence sentence of entity pair $(e_h, e_t)$.
Similar to entity embeddings, to obtain sentence embedding $\textbf{s}_n$, we apply a LogSumExp pooling over all the tokens in $s_n$: $\mathbf{s}_{n} = \log \sum_{h_l \in s_n} \exp \left( \mathbf{h}_l \right)$.
Intuitively, if $s_n$ is an evidence sentence of $(e_h, e_t)$, the tokens in $s_n$ would be relevant to the relation prediction, and should contribute more to $\textbf{c}_{h,t}$.
Hence, we use a bilinear function between context embedding $\textbf{c}_{h,t}$ and sentence embedding $\textbf{s}_n$ to measure the importance of sentence $s_n$ to entity pair $(e_h, e_t)$:
\begin{equation}
 \begin{aligned}
     \mathrm{P}\left(s_n|e_h, e_t \right) &= \sigma \left(\mathbf{s}_{n} \mW_v \textbf{c}_{h,t} + \vb_v\right), \\
 \end{aligned}
\label{eq:evi_bilinear}
\end{equation}
where $\mW_v$ and $\vb_v$ are learnable parameters. 

As an entity pair may have more than one evidence sentence, we use the binary cross entropy as the objective to train the evidence extraction model. 
\begin{gather}
    \begin{aligned}
    \mathcal{L}_{Evi} =& -\sum_{h \neq t, \text{NA} \notin \mathcal{P}^T_{h,t}} \sum_{s_n\in \mathcal{D}} y_n \cdot \mathrm{P}\left(s_n|e_h, e_t \right) + \\ & (1 - y_n) \cdot \log (1 - \mathrm{P}\left(s_n|e_h, e_t \right) ),
    \end{aligned}
\raisetag{15pt}
\end{gather}
where the evidence label $y_n$ is 1 when $s_n \in V_{h,t}$ and otherwise 0. If golden labels are not provided, we use several heuristic rules to construct silver labels instead. Details are introduced in Sec~\ref{sec:rule}.

Finally, we optimize our model by the combination of the relation extraction loss $\mathcal{L}_{RE}$ and evidence extraction loss $\mathcal{L}_{Evi}$:
\begin{equation}
    \mathcal{L} = \mathcal{L}_{RE} + \mathcal{L}_{Evi}. \label{eq:loss}
\end{equation}

\start{\edit{Efficiency Considerations}}
\edit{Compared to a previous method E2GRE \cite{E2GRE} that also extracts the evidence, \ours is significantly more efficient in both memory and training time for two reasons.
First, E2GRE learns $|\mathcal{R}|$ representations for each sentence. Namely, it makes evidence prediction for every (entity, entity, sentence, relation) tuple, which requires expensive computation especially when $|\mathcal{R}|$ is large (e.g., $|\mathcal{R}|$ = 96 in DocRED)}.
In contrast, we observe that most entity pairs only have one set of evidence across relations and thus predict only one set of evidence for each entity pair.

\edit{Second, E2GRE regards the evidence label of entity pairs with $r=\text{NA}$ as an empty set.
However, these entity pairs may still involve some relation beyond the pre-defined relation set $\mathcal{R}$, which also have their evidence sentences.}
Hence, we train the evidence extraction model only on entity pairs with at least one non-NA relation, which accounts for a small subset (e.g., 2.97\% in DocRED) of all the entity pairs.
\edit{Experiments show that \ours achieves better performances than E2GRE in both RE and evidence extraction while requiring only 30\% of its memory usage and 11\% of its runtime.} 

\edit{Furthermore, E2GRE does not utilize the evidence after extraction and relies heavily on the human annotation of evidence, which we will address in the following sections.}

\subsection{\edit{Fusion of Evidence in Inference}}
\label{sec:inference}
Suppose the extracted evidence sentences already contain all the information relevant to the relation, then there is no need to use the whole document for relation extraction. 
However, no system can perfectly extract the evidence without missing any sentences. 
Solely relying on the extracted evidence may miss important information in the document and lead to sub-optimal performance. 
Therefore, we combine the prediction results on both the original document and the extracted evidence, which can either be learned by our evidence classifier (Sec.~\ref{sec:joint}) or constructed by our heuristic rules (Sec.~\ref{sec:rule}) if evidence annotation is unavailable. 
\edit{Even without joint training, one may directly improve general (trained) DocRE models by applying our proposed inference process (noted as \ours (Rule)-Nojoint in Table \ref{tab:ablation}).
}

Specifically, as shown in Figure~\ref{fig:framework}, we first obtain a set of relation prediction scores $S_{h,t,r}^{(O)}$ from the original documents.
Then we construct a pseudo document $d'_{h,t}$ for each entity pair by concatenating the extracted evidence sentences $V'_{h,t}$ in the order they present in the original document.
The prediction score of the RE model on the pseudo document is noted as $S_{h,t,r}^{(E)}$. 
Finally, we fuse the results by aggregating the two sets of prediction sores through a blending layer \citep{Blending}:
\begin{equation}
    \mathrm{P}_{Fuse} \left(r|e_h, e_t\right) = \sigma( S^{(O)}_{h,t,r} + S^{(E)}_{h,t,r} - \tau ).
\end{equation}
We choose this design because it is simple and only includes one learnable parameter, $\tau$, alleviating over-fitting in the development set.
We optimize the parameter $\tau$ on the development set as follows:
\begin{gather}
    \begin{aligned}
    \mathcal{L}_{Fuse} &= -\sum_{d\in \mathcal{D}}
    \sum_{h \neq t}  \sum_{r\in \mathcal{R}}
    y_r \cdot \mathrm{P}_{Fuse} \left(r|e_h, e_t \right) +\\ 
    &(1 - y_r) \cdot 
    \log (1 - \mathrm{P}_{Fuse} \left(r|e_h, e_t \right) ),
    \end{aligned}
\raisetag{15pt}
\end{gather}
where $y_r = 1$ if relation $r$ holds between $(e_h, e_t)$ and $y_r = 0$ otherwise. 
Empirically, using other loss functions does not affect the performance much.

\subsection{Heuristic Evidence Label Construction}
\label{sec:rule}
In case that human annotation of evidence is not available, we design a set of heuristic rules to automatically construct silver labels for evidence extraction.
Then we train our joint model on the silver labels and directly use the silver labels as pseudo documents in inference.
The percentage of test samples covered by each rule is shown in Table~\ref{tab:breakdown}.

\smallskip
\start{Co-occur} If the head and tail entities co-occur in the same sentence (e.g., ``Load'' and ``the United States'' co-occur in the 1$^{st}$ sentence in Figure~\ref{fig:framework}), we use all the sentences they co-occur as evidence.

\smallskip
\start{Coref} If the proper-noun mentions of the head and tail entity do not co-occur, but their coreferential mentions co-occur (e.g., ``Hero of the Day'' and ``the album'', the co-reference of ``Load'' co-occur in the 10$^{th}$ sentence in Figure~\ref{fig:framework}), we use all the sentences where their coreferential mentions co-occur as evidence.
In practice, we directly apply a pre-trained coreference resolution model, HOI~\cite{hoi}, without fine-tuning on our dataset.

\smallskip
\start{Bridge} If the first two conditions are not met, but there exists a third bridge entity whose coreferential mention co-occurs with both head and tail (e.g., ``Load'' or its coreferential mention ``the album'' co-occurs with both ``the United States'' and ``Hero of the Day'' in Figure~\ref{fig:framework}), we take all the sentences where the bridge co-occurs with head or tail as the evidence. If there are more than one bridge entities, we choose the one with the highest frequency.
While this rule can be easily extended to multiple bridges, we empirically observe that capturing one bridge already leads to satisfying results.
\section{Experiments}

\begin{table*}[htb]
\centering
\scalebox{0.77}{
\begin{tabular}{lcccccc}
\toprule
\multirow{2}{*}{\bf Model} & \multicolumn{4}{c}{\bf Dev} & \multicolumn{2}{c}{\bf Test}  \\ 
\cmidrule(lr){2-5} \cmidrule(lr){6-7}
~ & \bf Ign F1 & \bf F1 & \bf Intra F1 & \bf Inter F1 & \bf Ign F1 & \bf F1 \\
\midrule
LSR-BERT$_{\text{base}}$ \citep{LSR} & 52.43 & 59.00 & 65.26 & 52.05 & 56.97 & 59.05 \\
GLRE-BERT$_{\text{base}}$ \citep{GLRE} & - & - & - & - & 55.40 & 57.40 \\
Reconstruct-BERT$_{\text{base}}$ \citep{Reconstruct} & 58.13 & 60.18 & - & - & 57.12 & 59.45 \\ 
GAIN-BERT$_{\text{base}}$ \citep{GAIN} & 59.14 & 61.22 & 67.10 & 53.90 & 59.00 & 61.24 \\ 
\midrule
BERT$_{\text{base}}$ \citep{finetune-bert} & - & 54.16 & 61.61 & 47.15 & - & 53.20 \\
BERT-Two-Step \citep{finetune-bert} & - & 54.42 & 61.80 & 47.28 & - & 53.92 \\
HIN-BERT$_{\text{base}}$ \citep{HIN} & 54.29 & 56.31 & - & - & 53.70 & 55.60 \\
E2GRE-BERT$_{\text{base}}$ \citep{E2GRE} & 55.22 & 58.72 & - & - & - & - \\
CorefBERT$_{\text{base}}$ \citep{CorefBERT} & 55.32 & 57.51 & - & - & 54.54 & 56.96 \\
ATLOP-BERT$_{\text{base}}$ \citep{ATLOP} & 59.11 $\pm$ 0.14$^\dagger$ & 61.01 $\pm$ 0.10$^\dagger$ & 67.26 $\pm$ 0.15$^\dagger$ & 53.20 $\pm$ 0.19$^\dagger$ & 59.31 & 61.30 \\
\midrule
\textbf{\ours(Rule)-BERT$_{\text{base}}$} & 60.36 $\pm$ 0.13 & 62.34 $\pm$ 0.08 & 68.40 $\pm$ 0.14 & 54.79 $\pm$ 0.13 & 60.23 & 62.21 \\
\textbf{\ours-BERT$_{\text{base}}$} & \textbf{60.51 $\pm$ 0.11} & \textbf{62.48 $\pm$ 0.13} & \textbf{68.47 $\pm$ 0.08} & \textbf{55.21 $\pm$ 0.21} & \textbf{60.42} & \textbf{62.47} \\
\midrule
\midrule
RoBERTa$_{\text{large}}$~\cite{CorefBERT}& 57.14 & 59.22 & - & - & 57.51 & 59.62 \\
CorefRoBERTa$_{\text{large}}$~\cite{CorefBERT}& 57.35 & 59.43 & - & - & 57.90 & 60.25\\
E2GRE-RoBERTa$_{\text{large}}$~\cite{E2GRE}& 59.55 & 62.91 & - & - & 60.29 & 62.51 \\
GAIN-BERT$_{\text{large}}$~\cite{GAIN}& 60.87 & 63.09 & - & - & 60.31 & 62.76\\
ATLOP-RoBERTa$_{\text{large}}$~\cite{ATLOP}& 61.30 $\pm$ 0.22$^\dagger$ & 63.15 $\pm$ 0.21$^\dagger$ & 69.61 $\pm$ 0.25$^\dagger$ & 55.01 $\pm$ 0.18$^\dagger$ & 61.39 & 63.40 \\
\midrule
\textbf{\ours(Rule)-RoBERTa$_{\text{large}}$} & 61.73 $\pm$ 0.07 & 63.91 $\pm$ 0.07 & 69.99 $\pm$ 0.09 & 56.27 $\pm$ 0.11 & 61.93 & 64.12 \\
\textbf{\ours-RoBERTa$_{\text{large}}$} & \textbf{62.34 $\pm$ 0.14} & \textbf{64.27 $\pm$ 0.10} & \textbf{70.36 $\pm$ 0.07} & \textbf{56.53 $\pm$ 0.15} & \textbf{62.85} & \textbf{64.79} \\
\bottomrule
\end{tabular}
}
\upv
\caption{Relation extraction results on DocRED.
We report the mean and standard deviation on the development set by conducting 5 runs with different random seeds.
We report the official test score of the best checkpoint on the development set.
Results with $\dagger$ are based on our implementation. Others are reported in their original papers. 
We separate graph-based and transformer-based methods into two groups.
}
\label{table:results}
\downv
\end{table*}

\begin{table}[htb]
\centering
\scalebox{0.67}{  
    \begin{tabular}{lcc}
         \toprule
         \bf Model & \bf CDR & \bf GDA \\
         \midrule
         LSR-BERT$_{\text{base}}$~\cite{LSR} & 64.8 & 82.2\\
         SciBERT$_{\text{base}}$~\cite{ATLOP} & 65.1 $\pm$ 0.6 & 82.5 $\pm$ 0.3 \\
         DHG-BERT$_{\text{base}}$~\cite{DHG} & 65.9 & 83.1 \\
         GLRE-SciBERT$_{\text{base}}$~\cite{GLRE}& 68.5 & - \\
         ATLOP-SciBERT$_{\text{base}}$~\cite{ATLOP}& 69.4 $\pm$ 1.1 & 83.9 $\pm$ 0.2 \\
         \midrule
         \textbf{\ours (Rule)-SciBERT$_{\text{base}}$}& \textbf{70.63} $\pm$ 0.49 & \textbf{84.54} $\pm$ 0.22 \\
         \bottomrule
    \end{tabular}}
    \upv
    \caption{Relation extraction results on CDR and GDA.}
    \label{tab::bio}
    \downv
\end{table}

\subsection{Experiment Setup}
\label{sec:setup}
\start{Datasets}
We evaluate the effectiveness of \ours on three datasets: DocRED~\cite{DocRED-paper}, CDR~\cite{CDR} and GDA~\cite{GDA}, where DocRED is the only dataset that provides evidence labels as part of the annotation. The details of the datasets are listed in Appendix~\ref{app:data}.

\smallskip
\start{Implementation Details}
Our model is implemented based on PyTorch and Huggingface's Transformers \cite{wolf2019huggingface}. We use cased-BERT$_{\text{base}}$ \cite{BERT} and RoBERTa$_{\text{large}}$ as the base encoders and optimize our model using AdamW with learning rate 5e-5 for the encoder and $1e-4$ for other parameters. We adopt a linear warmup for the first 6\% steps. The batch size (number of documents per batch) is set to 4 and the ratio between relation extraction and evidence extraction losses is set to 0.1. We perform early stopping based on the F1 score on the development set, with a maximum of 30 epochs. Our BERT$_{\text{base}}$ models are trained with one GTX 1080 Ti GPU and RoBERTa$_{\text{large}}$ models with one RTX A6000 GPU.

\smallskip
\start{Evaluation Metrics}
Following prior studies~\cite{DocRED-paper}, we use \textbf{F1} and \textbf{Ign F1} as the main evaluation metrics for relation extraction, where \textbf{Ign F1} measures the F1 score excluding the relations shared by the training and development/test set.
We also report \textbf{Intra F1} and \textbf{Inter F1}, where the former measures the performance on the co-occurred (intra-sentence) entity pairs and the latter evaluates the inter-sentence entity pairs where none of their proper-noun mentions co-occurs.
For evidence extraction, we compute the F1 score (denoted as \textbf{Evi F1}) and further introduce \textbf{PosEvi F1}, which measures the F1 score of evidence only on positive entity pairs (i.e., those with non-NA relations).

\subsection{Main Results}
We compare our methods with both \emph{Graph-based methods} and \emph{transformer-based methods}. Graph-based methods explicitly perform inference on document-level graphs. Transformer-based methods, including \ours, implicitly capture the long-distance token dependencies via transformers.
Noted that \ours is trained on gold labels and leverages the evidence extracted by our model in inference. \ours (Rule) is trained on silver evidence labels constructed by rules and also leverages them in inference. 

\smallskip
\start{Relation Extraction Results}
Tables~\ref{table:results} and \ref{tab::bio} show that \ours outperforms the DocRE baseline methods in all datasets.
Our improvement is especially large on Inter F1 (e.g., 1.21/2.01 Intra/Inter F1 compared to ATLOP-BERT$_{\text{base}}$).
We hypothesize that the bottleneck of inter-sentence pairs is to locate the relevant context, which often spreads through the whole document. \ours learns to capture important sentences in training and focuses more on these important sentences in inference.

Among the baselines, the Inter F1 of GAIN is 0.70 higher than ATLOP while the Intra F1 of ATLOP is 0.16 higher than GAIN, indicating that document-level graphs may be effective in multi-hop reasoning.
Although \ours does not involve explicit multi-hop reasoning modules, it still notably outperforms graph-based models in Inter F1.

Finally, \ours (Rule) also outperforms all the baselines in both DocRED and the two biomedical datasets which do not have evidence annotation.
The improvement on DocRED and CDR is much larger than that on GDA. We hypothesize that it is because more than 85\% relations in GDA are intra-sentence ones, making it trivial even for the single RE model to focus on these sentences.

\begin{table}[t]
\centering
\scalebox{0.75}{
    \begin{tabular}{lccc}
         \toprule
         \textbf{Model} & \bf Dev Evi F1 & \bf Test Evi F1  \\
         \midrule
         E2GRE-BERT$_{\text{base}}$ & 47.14 & 48.35 \\
         \textbf{\ours-BERT$_{\text{base}}$} & \textbf{50.71} & \textbf{51.27} \\
         \midrule
         E2GRE-RoBERTa$_{\text{large}}$ & 51.11 & 50.50 \\
         \textbf{\ours-RoBERTa$_{\text{large}}$} & \textbf{52.54} & \textbf{53.01} \\
         \bottomrule
    \end{tabular}}
    \upv
    \caption{Evidence extraction results on DocRED. We compare \ours with E2GRE \citep{E2GRE}.}
    \label{tab:evidence}
    \downv
\end{table}

\smallskip
\start{Evidence Extraction Results} 
To our knowledge, E2GRE is the only method that has reported their evidence extraction result.
The results in Table~\ref{tab:evidence} indicate that \ours outperforms E2GRE significantly (e.g., by 3.57 Dev Evi F1 under BERT$_{\text{base}}$).
The results show that it may be sufficient to train the evidence classifier only on pairs with $r\in\mathcal{R}$ and over each (entity, entity, sentence) tuple instead of (entity, entity, sentence, relation) as in E2GRE.

Our ablation studies in Table~\ref{tab:evi_ablation} show that our three heuristic rules, denoted as \textbf{Rules (ours)}, already capture most of the evidence for positive entity pairs. The high quality of silver labels explains why our model can perform well using silver labels only.
Furthermore, training the RE model and evidence extraction model separately (denoted as \textbf{NoJoint}) results in a sharp performance drop. 
As the relation and evidence classifiers share the same base encoder, discarding the relation classifier will result in insufficient training of the base encoder and harm the performance.

\begin{table}[t]
\centering
\scalebox{0.75}{
    \begin{tabular}{lccc}
         \toprule
           & Rules (ours) & \ours-BERT$_{\text{base}}$ &  NoJoint \\
         \midrule
          \bf PosEvi F1 & 77.43 & \textbf{80.33} & 51.13 \\
         \bottomrule
    \end{tabular}}
    \upv
    \caption{Ablation study for evidence extraction.}
    \label{tab:evi_ablation}
    \downv
\end{table}

\subsection{Performance Analysis}

\start{Ablation Study} Table \ref{tab:ablation} shows the ablation studies that analyzes the utility of each module in \ours.
We observe that \textbf{NoJoint} leads to sharp performance drop in DocRE.
Besides, \textbf{\ours (Rule)-Nojoint} achieves significant ``free gains'' (0.90/1.08 Ign F1/F1) by simply fusing the evidence constructed by rules in the inference of ATLOP. In principle, this inference process can be applied to general DocRE models.

\begin{table}[!t]
\centering
\scalebox{0.72}{
    \begin{tabular}{lcccc}
         \toprule
         \textbf{Ablation} & \bf Ign F1 & \bf F1 & \bf Intra F1 & \bf Inter F1 \\
         \midrule
         \ours-BERT$_{\text{base}}$ & \textbf{60.51} & \textbf{62.48} & \textbf{68.47} & \textbf{55.21} \\
            \quad NoJoint & 59.98 & 62.03 & 68.51 & 54.10 \\
            \quad NoPseudo & 59.70 & 61.53 & 67.55 & 54.01 \\
            \quad NoOrigDoc & 58.47 & 60.44 & 66.24 & 53.23 \\
            \quad NoBlending & 58.93 & 61.46 & 67.33 & 54.37 \\
            \quad FinetuneOnEvi & 60.11 & 62.29 & 68.13 & 54.84 \\
        \midrule
        \ours(Rule)-BERT$_{\text{base}}$ & \textbf{60.36} & \textbf{62.34} & \textbf{68.40} & \textbf{54.79} \\
            \quad NoJoint & 60.01 & 62.09 & 68.21 & 54.34 \\
         \bottomrule
    \end{tabular}}
    \upv
    \caption{Ablation study of \ours on DocRED.}
    \label{tab:ablation}
    \downv
\end{table}

We also remove the pseudo document (constructed from the extracted evidence) and the original document separately, denoted as \textbf{NoPseudo} and \textbf{NoOrigDoc}, respectively.
We observe that removing either source will lead to performance drops. Also, the drop of Inter F1 is much larger than Intra F1 for \textbf{NoPseudo},
indicating that our inference process is effective for inter-sentence pairs where the evidence may not be consecutive.

As for \textbf{NoBlending}, we remove the blending layer and simply take the union of the two sets of results.
The sharp drop of performance indicates
the blending layer can successfully learn a dynamic threshold to combine the prediction results.

Finally, we further finetune the RE model on ground truth evidence before feeding it the extracted evidence (denoted as \textbf{FinetuneOnEvi}) but the performance is not improved, probably because the encoded entity representations in evidence and original documents are already highly similar.

\begin{table}[!t]
\centering
\scalebox{0.8}{
    \begin{tabular}{lcccc}
         \toprule
          & \bf Co-occur & \bf Coref & \bf Bridge & \bf Total \\
         \midrule
          Count & 6711 & 984 & 3212 & 10,907 \\ %
          Percent & 54.46\% & 7.99\% & 26.07\% & 88.52\% \\ %
         \bottomrule
    \end{tabular}}
    \upv
    \caption{Statistics of the 12,323 relations in the DocRED development set.}
    \label{tab:breakdown}
    \downv
\end{table}

\begin{figure}[t]
\centering
\scalebox{0.95}{
    \includegraphics[width=\linewidth]{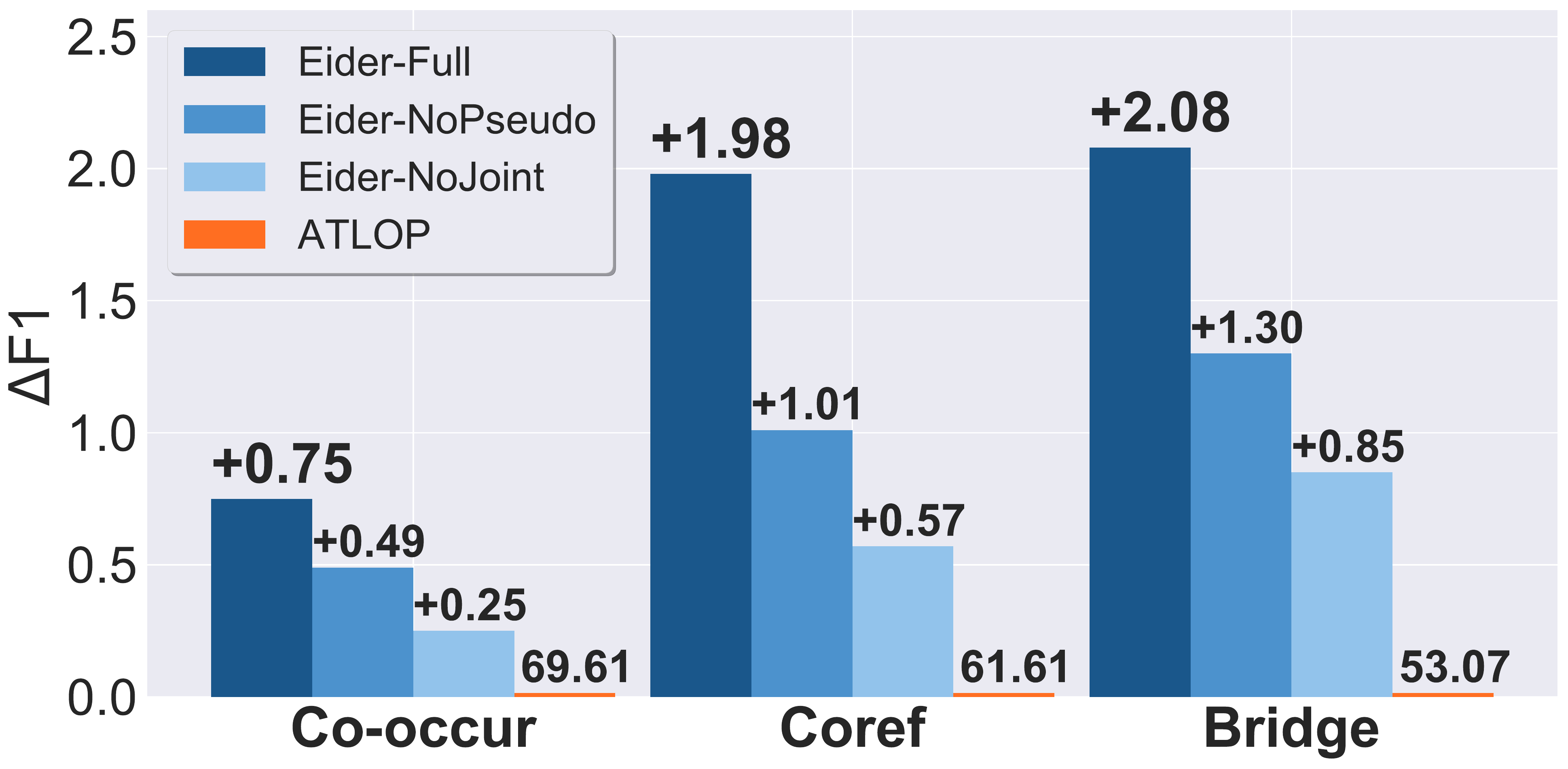}
}
    \upv
    \caption{Performance gains in F1 by relation categories. The gains are relative to the second best baseline (ATLOP-RoBERTa$_{\text{large}}$).}
    \label{fig:breakdown}
    \downv
    \vspace{-0.17cm}
\end{figure}

\begin{table}[b]
\centering
\scalebox{0.78}{
    \begin{tabular}{lcc}
         \toprule
         \textbf{Model} & \bf Memory & \bf Training time \\
         \midrule
            ATLOP-BERT$_{\text{base}}$ & 9,139 MB & 5.19 it/s \\
            E2GRE-BERT$_{\text{base}}$ & 36,182 MB & 0.53 it/s \\
            \ours-BERT$_{\text{base}}$ & 10,933 MB & 4.92 it/s \\
         \bottomrule
    \end{tabular}}
    \upv
    \caption{Training time and memory usage on DocRED.}
    \label{tab:efficiency}
    \downv
\end{table}

\begin{table*}[ht]
\centering
        \resizebox{2.08\columnwidth}{!}{
        \begin{tabular}{p{20cm}}
            \toprule
            \textbf{Ground Truth Relation: \colorG{Located in}} 
            \quad
            \textbf{Ground Truth Evidence Sentence(s)}: [1, 2]
            \quad
            \textbf{Extracted Evidence Sentence(s)}: \colorbox{LightYellow}{[1, 2]}
            \\
             \textbf{Document}: \colorbox{LightYellow}{[1]} The \colorR{Portland Golf Club} is a private golf club in the northwest United States , in suburban Portland, Oregon.
            \colorbox{LightYellow}{[2]} It is located in the unincorporated Raleigh Hills area of eastern \colorB{Washington County}, southwest of downtown Portland and east of Beaverton.
            [3] The club was established in the winter of 1914, when a group of nine businessmen assembled to form a new club after leaving their respective clubs \textbf{...}
             \\ 
             \textbf{Final Prediction}: \colorG{Located in} 
             \quad\quad
             \textbf{Prediction on Orig. Doc}: \colorG{Located in} 
             \quad\quad
             \textbf{Prediction on Extracted Evidences}: \colorG{Located in}
             \\
             
            \midrule
            
            \textbf{Ground Truth Relation: \colorG{Characters}}
            \hskip0.5em
            \textbf{Ground Truth Evidence Sentence(s)}: [1, 3]
            \quad
            \textbf{Extracted Evidence Sentence(s)}: \colorbox{LightYellow}{[1, 3]} 
            \\
             \textbf{Document}: \colorbox{LightYellow}{[1]} King Louie is a fictional character introduced in Walt Disney's 1967 animated musical film, \colorR{The Jungle Book}. [2] Unlike the majority of the adapted characters in the film, Louie was not featured in Rudyard Kipling's original works. \colorbox{LightYellow}{[3]} King Louie was portrayed as an orangutan who was the leader of the other jungle primates, and who attempted to gain knowledge of fire from \colorB{Mowgli}, \textbf{...}
             \\ 
             \textbf{Final Prediction}: \colorG{Characters}
             \quad\quad
             \textbf{Prediction on Orig. Doc}: NA
             \quad\quad\quad\quad\quad
             \textbf{Prediction on Extracted Evidences}: \colorG{Characters}
             \\
             
            \midrule
            \textbf{Ground Truth Relation: \colorG{Inception}} 
            \quad
            \textbf{Ground Truth Evidence Sentence(s)}: [5, 6]
            \quad\quad
            \textbf{Extracted Evidence Sentence(s)}: \colorbox{LightYellow}{[5]}
            \\
             \textbf{Document}: [1] Oleg Tinkov (born 25 December 1967 ) is a Russian entrepreneur and cycling sponsor. \textbf{...} \colorbox{LightYellow}{[5]} Tinkoff is the founder and chairman of the \colorR{Tinkoff Bank} board of directors (until 2015 it was called Tinkoff Credit Systems). [6] The bank was founded in \colorB{2007} and as of December 1, 2016, it is ranked 45 in terms of assets and 33 for equity among Russian banks. \textbf{...}
             \\ 
             \textbf{Final Prediction}: \colorG{Inception}
             \quad\quad\quad
             \textbf{Prediction on Orig. Doc}: \colorG{Inception}
             \quad\quad\quad\quad\quad
             \textbf{Prediction on Extracted Evidences}: NA
             \\
             
            \bottomrule
        \end{tabular}
        }
    \upv
    \caption{Case studies of our proposed framework \ours. We use red, blue and green to color the \colorR{head entity}, \colorB{tail entity} and \colorG{relation}, respectively. The indices of \colorbox{LightYellow}{extracted evidence sentences}are highlighted with yellow.}
    \label{tab:case}
    \downv
    \end{table*}

\smallskip
\start{Performance Breakdown}
To further analyze the performance of \ours on different types of entity pairs, we categorize the relations into three categories based on our three heuristic rules in Sec.~\ref{sec:rule}: \textit{Co-occur}, \textit{Coref} and \textit{Bridge}.
The number and percentage of relations covered by each rule are listed in Table~\ref{tab:breakdown}. We can see that the three categories cover over 88\% of the relations in the development set.
The results on each category are shown in Figure~\ref{fig:breakdown}. We can see that our full model has the best performance in all three categories and our ablations also outperform ATLOP.
For all our methods, the improvements over ATLOP is \textit{Bridge} $>$ \textit{Coref} $\gg$ \textit{Co-occur}. This reveals that both modules mainly improve the model's reasoning ability from multiple sentences, either by coreference reasoning or by multi-hop reasoning over a third entity.

\smallskip
\start{Efficiency Comparison}
We benchmark the time and memory usage of \ours on an RTX A6000 GPU.
Table~\ref{tab:efficiency} shows that our joint model incurs only \textasciitilde5\% training time and \textasciitilde14\% GPU memory overhead.
Experiments also show that \ours can be trained on a single consumer GPU (e.g., an 11GB GTX 1080 Ti) but E2GRE is not able to.

\subsection{Case Studies}

Table~\ref{tab:case} shows a few examples of \ours. 
Detailed statistics and error analysis are provided in Appendix \ref{app:err_ana}.
In the first example, the head entity is mentioned in the first sentence and the tail entity appears in the second. 
We can see that \ours correctly extracts these sentences as evidence. 
Since the evidence sentences are consecutive, the predictions on both the original document and the evidence sentences are correct. 
In the second example, 
the prediction using only the original document is incorrect, possibly because the ``King Louie'' in the $1^{st}$ and $3^{rd}$ sentences are so far away from each other that the model fails to recognize them as coreference. Hence, it fails to distinguish ``King Louie'' as a bridge entity and wrongly predicts ``NA''.
Instead, these two sentences are consecutive in the extracted evidence, making it easier for the model to find the bridge. 
In the last example, the $6^{th}$ sentence is missing in the extracted evidence, 
so the extracted evidence does not contain enough information to predict the relation. However, the prediction on the original document is correct, leading to the correct final result.

\section{Related Work}
\start{Relation Extraction} 
Previous research efforts on relation extraction mainly concentrate on predicting relations within a sentence \citep{cai2016bidirectional,zhang2018attention,zhang2019long,zhang2020relation}. 
Despite their effectiveness, in the real world, certain relations can only be inferred from multiple sentences. 
Consequently, recent studies~\citep{quirk-poon-17-distant,peng-etal-17-cross,DocRED-paper}
started to work on document-level relation extraction (DocRE).

\smallskip
\start{Graph-based DocRE} 
Graph-based DocRE methods generally construct a graph with mentions, entities, sentences, or documents as the nodes, and infer the relations by reasoning on this graph.
\citet{GAIN} performs multi-hop reasoning on both a mention-level graph and an entity-level graph.
\citet{Reconstruct} extracts a reasoning path for each relation and encourages the model to reconstruct the path during training.
\citet{SIRE} separately deals with intra- and inter-sentential entity pairs and performs multi-hop reasoning on a mention-level graph for inter-sentential entity pairs.
However, the extracted graph may omit some important information in the text. Complicated operations on the graphs may also hinder the model from capturing the text structure.

\smallskip
\start{Transformer-based DocRE} 
Another line of studies model cross-sentence relations by implicitly capturing the long-distance token dependencies via the transformer \cite{transformer}.
\citet{ATLOP} uses attention in the transformers to extract useful context and adopts an adaptive threshold for each entity pair.
\citet{DocuNet} views DocRE as a semantic segmentation task over the entity matrix and applies a U-Net to capture the correlations between relations.
\edit{\citet{E2GRE} guides DocRE by extracting evidence but does not leverage them after extraction. It also highly relies on evidence annotations and suffers from massive runtime and memory overhead.}
\citet{3Sent} predicts on only a few sentences selected by rules, which may miss important information and does not show consistent improvements.
\edit{In comparison, we design a lightweight evidence extraction model that is significantly more efficient than \citet{E2GRE} and can improve DocRE even trained on silver labels.
\ours also fuses the extracted evidence in inference, putting more attention to the important sentences without information loss.}

\section{Conclusion}
In this work, we propose \ours, an \textbf{e}v\textbf{id}ence-\textbf{e}nhanced \textbf{R}E framework, which improves DocRE by joint relation and evidence extraction and fusion of extracted evidence in inference. 
In training, the RE and evidence extraction model provide additional training signals for each other and mutually enhance each other. 
The joint model is efficient in time and memory and does not rely heavily on the human annotation of evidence.
During inference, the prediction results on both the original document and the extracted evidence are combined, which encourages the model to focus on the important sentences while reducing information loss.
Experiment results demonstrate that \ours significantly outperforms existing methods on three public datasets (DocRED, CDR, and GDA), especially on inter-sentence relations. 

\section*{Acknowledgements}
Research was supported in part by US DARPA KAIROS Program No. FA8750-19-2-1004, SocialSim Program No.  W911NF-17-C-0099, and INCAS Program No. HR001121C0165, National Science Foundation IIS-19-56151, IIS-17-41317, and IIS 17-04532, and the Molecule Maker Lab Institute: An AI Research Institutes program supported by NSF under Award No. 2019897. Any opinions, findings, and conclusions or recommendations expressed herein are those of the authors and do not necessarily represent the views, either expressed or implied, of DARPA or the U.S. Government.

\bibliography{custom}
\bibliographystyle{acl_natbib}
\newpage
\clearpage
\appendix

\section{Appendices}
\subsection{Dataset Statistics}
\label{app:data}
Our model is evaluated on three benchmark datasets, where the statistics are shown in Table~\ref{tab:dataset}:

\textbf{DocRED~\cite{DocRED-paper}} is a large human-annotated document-level RE dataset constructed from Wikipedia.
In the training set, around 97.03\% entity pairs do not hold any explicit relations.
In our experiments, the performance on the test set is validated through the Leader board\footnote{Results can be found at \url{https://competitions.codalab.org/competitions/20717}.}.

\textbf{CDR~\cite{CDR}} is a biomedical relation extraction dataset consisting of 1,500 PubMed abstracts.
The only two entity types are chemicals and diseases and the only non-NA relation is the causal relation between chemicals and disease concepts.

\textbf{GDA~\cite{GDA}} contains 30,192 MEDLINE abstracts. It is also a biomedical dataset with two entity types only: diseases and genes, and one non-NA relation type only: the interactions between disease concepts and genes.

\begin{table}[hb]
\centering
\scalebox{0.85}{
    \begin{tabular}{lccc}
         \toprule
          \bf Statistics & DocRED & CDR & GDA \\
         \midrule
         \# Train & 3053 & 500 & 23353 \\
         \# Dev & 1000 & 500 & 5839 \\
         \# Test & 1000 & 500 & 1000 \\
         \# Relation types & 97 & 2 & 2 \\
         \# Avg.\# entities per Doc & 19.5 & 7.6 & 5.4 \\
         \# Avg.\# sentences per Doc & 8.0 & 9.7 & 10.2 \\
         Percent of Intra Rel & 54.2 & 75.7 & 84.7 \\
         \bottomrule
    \end{tabular}}
    \upv
    \caption{Statistics of the datasets in experiments. The percentage of intra-sentence relations is calculated from the development set of DocRED and calculated from the test set of CDR and GDA.}
    \label{tab:dataset}
    \downv
\end{table}

\subsection{Error Analysis of \ours}
\label{app:err_ana}

\begin{table}[t]
\centering
\resizebox{.85\linewidth}{!}{%
\begin{tabular}{lllc} & & \multicolumn{2}{c}{\bf Ground Truth} \\ 
\cline{2-4} 
\multicolumn{1}{l|}{\multirow{4}{*}{\rotatebox{90}{\bf Prediction}}}
& \multicolumn{1}{l|}{} 
& \multicolumn{1}{l|}{$r\in\mathcal{R}$} 
& \multicolumn{1}{c|}{NA} \\ 
\cline{2-4} 
 \multicolumn{1}{l|}{} 
& \multicolumn{1}{l|}{$r\in\mathcal{R}$ (Correct)} 
& \multicolumn{1}{l|}{7,696 (\checkmark)} 
& \multicolumn{1}{c|}{\multirow{2}{*}{3,613 (\xmark)}} \\ \cline{2-3}
\multicolumn{1}{l|}{} 
& \multicolumn{1}{l|}{$r\in\mathcal{R}$ (Wrong)}   
& \multicolumn{1}{l|}{287 (\xmark)}      
& \multicolumn{1}{c|}{} \\ 
\cline{2-4} 
\multicolumn{1}{l|}{} 
& \multicolumn{1}{l|}{NA}                 
& \multicolumn{1}{l|}{4,340 (\xmark)}     
& \multicolumn{1}{c|}{380,854 (\checkmark)} \\ 
\cline{2-4} 
\end{tabular}%
}
    \caption{Statistics of one run of  \ours-RoBERTa$_{\text{large}}$. ``$r\in\mathcal{R}$'' means non-NA relations. We use ``\checkmark'' and ``\xmark'' to denote correct and wrong predictions, respectively. For example, we have 4,340 wrong predictions where the ground truth is some $r\in\mathcal{R}$ but the prediction is NA.}
    \label{tab:confusion}
\end{table}

\begin{table}[htb]
\centering
\scalebox{0.86}{
    \begin{tabular}{lc}
         \toprule
           Reason & Count \\
         \midrule
          Labeling Mistakes & 18 \\
          Fail in Commonsense Reasoning & 8 \\
          Fail in Coreferential Reasoning & 6 \\
          Fail in Multi-hop Reasoning & 4 \\
          Fail in Surface-name Reasoning & 3 \\
          Wrong Evidence Extraction & 1 \\
          Others & 10 \\
         \bottomrule
    \end{tabular}}
    \upv
    \caption{Error types of \ours in 50 randomly sampled error cases in DocRED. Where ``Labeling Mistakes'' means our model predicts correctly but the annotation is wrong.}
    \label{tab:error_type}
    \downv
\end{table}

\begin{table*}[t]
\centering
        \resizebox{2.1\columnwidth}{!}{
        \begin{tabular}{p{22cm}}
            \toprule
            \textit{\textbf{Error Type 1}: Labeling Mistakes}\\
            \textbf{Ground Truth Relation: \colorG{Country} (\xmark)} 
            \quad\quad
            \textbf{Ground Truth Evidence Sentence(s)}: [1, 4, 5, 7]
            \quad\quad
            \textbf{Extracted Evidence Sentence(s)}: \colorbox{LightYellow}{[5, 7]}
            \\
             \textbf{Document}: 
             [1] Westmere is a hamlet in the town of Guilderland, Albany County, New York. 
             [4] It is a suburb of the neighboring city of Albany. 
             \colorbox{LightYellow}{[5]} \colorB{U.S. Route 20} (Western Avenue) bisects the community and is the major thoroughfare and main street. 
             \textbf{...}
             \colorbox{LightYellow}{[7]} Crossgates Mall, the \colorR{Capital District}'s largest shopping mall, is in Westmere's northeastern corner.
             \\ 
             \textbf{Final Prediction}: NA 
             \quad\quad\quad\quad\quad\quad\quad\quad\quad
             \textbf{Prediction on Orig. Doc}: NA 
             \quad\quad\quad\quad\quad\quad\quad\quad\quad\quad
             \textbf{Prediction on Extracted Evidences}: NA
             \\

            \midrule

            \textit{\textbf{Error Type 2}: Fail in Commonsense Reasoning}\\
            \textbf{Ground Truth Relation: \colorG{Located in}} 
            \quad\quad\quad
            \textbf{Ground Truth Evidence Sentence(s)}: [1, 5]
            \quad\quad\quad\quad
            \textbf{Extracted Evidence Sentence(s)}: \colorbox{LightYellow}{[1, 5]} 
            \\
             \textbf{Document}: 
             \colorbox{LightYellow}{[1]} Oakland / Troy Airport is a county-owned public-use airport located east of the central business district of Troy, a city in Oakland County, \colorB{Michigan}, United States.
             [2] It is included in the Federal Aviation Administration (FAA) National Plan of Integrated Airport Systems for 2017–2021, in which it is categorized as a regional reliever airport facility.
             \textbf{...}
             \colorbox{LightYellow}{[5]} It is located between Maple Road and 14 Mile Road and Coolidge Highway and \colorR{Crooks Road}.
             [6] \textbf{...}
             \\ 
             \textbf{Final Prediction}: NA (\xmark)
             \quad\quad\quad\quad\quad\quad\quad\quad
             \textbf{Prediction on Orig. Doc}: NA (\xmark)
             \quad\quad\quad\quad\quad\quad
             \textbf{Prediction on Extracted Evidences}: NA (\xmark)
             \\
             \\
            \textbf{Ground Truth Relation: \colorG{Religion}} 
            \quad\quad\quad\quad
            \textbf{Ground Truth Evidence Sentence(s)}: [1, 6]
            \quad\quad\quad
            \textbf{Extracted Evidence Sentence(s)}: \colorbox{LightYellow}{[1, 2, 6]} 
            \\
             \textbf{Document}: 
            \colorbox{LightYellow}{[1]} Marcial Maciel Degollado (March 10, 1920 – January 30, 2008) was a Mexican \colorB{Catholic} priest who founded the Legion of Christ and the Regnum Christi movement, serving as general director of the legion from 1941 to 2005.
            \colorbox{LightYellow}{[2]} Throughout most of his career, he was respected within the church as ``the greatest fundraiser of the modern \colorB{Roman Catholic} church'' and as a prolific recruiter of new seminarians.
            \textbf{...}
            \colorbox{LightYellow}{[6]} In 2006 Pope \colorR{Benedict XVI} removed Maciel from active ministry based on the results of an investigation that he had started while head of the Congregation for the Doctrine of the Faith, before his election as Pope in April 2005.
            \\
            \textbf{Final Prediction}: NA (\xmark)
            \quad\quad\quad\quad\quad\quad\quad\quad
            \textbf{Prediction on Orig. Doc}: NA (\xmark)
            \quad\quad\quad\quad\quad\quad
            \textbf{Prediction on Extracted Evidences}: \colorG{Religion}
            \\
            \midrule

            \textit{\textbf{Error Type 3}: Fail in Coreferential Reasoning}\\
            \textbf{Ground Truth Relation}: NA
            \quad\quad\quad\quad\quad\quad\quad
            \textbf{Ground Truth Evidence Sentence(s)}: []
            \quad\quad\quad\quad\quad
            \textbf{Extracted Evidence Sentence(s)}: \colorbox{LightYellow}{[1]}
            \\
             \textbf{Document}: 
             \colorbox{LightYellow}{[1]} \colorB{Manon Balletti} (1740–1776) was the daughter of Italian actors performing in France and lover of the famous womanizer \colorR{Giacomo Casanova}.
             [2] She was ten years old when she first met him; she happened to be the daughter of Silvia Balletti, an actress of the Comédie Italienne company and younger sister of \colorR{Casanova}'s closest friend.
             \textbf{...}
             \\ 
             \textbf{Final Prediction}: \colorG{Child} (\xmark)
             \quad\quad\quad\quad\quad\quad\quad
             \textbf{Prediction on Orig. Doc}:  \colorG{Child} (\xmark)
             \quad\quad\quad\quad
             \textbf{Prediction on Extracted Evidences}: \colorG{Child} (\xmark)
             \\
             
            \midrule
            \textit{\textbf{Error Type 4}: Fail in Multi-hop Reasoning}\\
            \textbf{Ground Truth Relation: \colorG{Educated at}} \quad\quad\quad
            \textbf{Ground Truth Evidence Sentence(s)}: [4]
            \quad\quad\quad\quad\quad
            \textbf{Extracted Evidence Sentence(s)}: \colorbox{LightYellow}{[4]}
            \\
             \textbf{Document}: 
             [1] Ronald Leonard is an American cellist.
             [2] He has had a distinguished career as a soloist, chamber musician, principal cellist and teacher.
             \textbf{...}
             \colorbox{LightYellow}{[4]} He was a winner of the Walter Naumburg Competition while a student at the \colorB{Curtis Institute of Music}, where he studied with Leonard Rose and \colorR{Orlando Cole}.
             \textbf{...}
             \\ 
             \textbf{Final Prediction}: NA (\xmark)
             \quad\quad\quad\quad\quad\quad\quad\quad
             \textbf{Prediction on Orig. Doc}: NA (\xmark)
             \quad\quad\quad\quad\quad\quad
             \textbf{Prediction on Extracted Evidences}: NA (\xmark)
             \\
             
            \midrule
            \textit{\textbf{Error Type 5}: Fail in Surface-name Reasoning}\\
            \textbf{Ground Truth Relation: \colorG{Country}} 
            \quad\quad\quad\quad
            \textbf{Ground Truth Evidence Sentence(s)}: []
            \quad\quad\quad\quad\quad
            \textbf{Extracted Evidence Sentence(s)}: \colorbox{LightYellow}{[1, 4]} 
            \\
             \textbf{Document}: 
             \colorbox{LightYellow}{[1]} A Route Army was a type of military organization during the Chinese Republic, and usually exercised command over two or more corps or a large number of divisions or independent brigades.
             [2] It was a common formation in \colorR{China} prior to the Second Sino-Japanese War but was discarded as a formation type by the National Revolutionary Army after 1938 (other than the 8th Route Army), in favor of the Group Army.
             [3] Some of the more famous of the Route Armies were:
             \colorbox{LightYellow}{[4]} 8th Route Army: Communist guerrilla force in \colorB{North China}. \textbf{...}
             \\ 
             \textbf{Final Prediction}: NA (\xmark)
             \quad\quad\quad\quad\quad\quad\quad\quad
             \textbf{Prediction on Orig. Doc}: NA (\xmark)
             \quad\quad\quad\quad\quad\quad
             \textbf{Prediction on Extracted Evidences}: NA (\xmark)
             \\
             
            \bottomrule
        \end{tabular}
        }
    \upv
    \caption{Examples for the four most common error types. We use red, blue and green to color the \colorR{head entity}, \colorB{tail entity} and \colorG{relation}, respectively. The indices of  \colorbox{LightYellow}{extracted evidence sentences}are highlighted with yellow.}
    \label{tab:error_case}
    \downv
    \end{table*}

The detailed statistics of the predictions of our model are listed in Table~\ref{tab:confusion}. Among all the errors, the majority is because the model wrongly predicts the non-NA relations (i.e., $r\in\mathcal{R}$) as ``NA'' or predicts ``NA'' as some non-NA relations. Only $\frac{287}{287+4340+3613}=3.48\%$ of the errors result from wrongly taking some non-NA relation as another.

To check the exact reason why our model makes these errors, we randomly select 50 cases from DocRED where our model predicts wrongly. We summarize the error types in Table~\ref{tab:error_type} and provide one or two examples for each of the common error types in Table~\ref{tab:error_case}.

Our analysis shows that 18 out of 50 ``error cases'' are actually correct. It suggests that labeling mistakes are still prevalent in the DocRED dataset. We show an example under \textbf{\textit{``Error Type 1''}} in Table~\ref{tab:error_case}. The annotator wrongly labels \textit{``U.S. Route 20''}, a highway, as the country of \textit{``Capital District''}.

Another common error type is \textit{\textbf{``Error Type 2''}: failing in commonsense reasoning}.
These error examples normally require commonsense knowledge of the related entities that does not explicitly present in the document.
In the first case, the document shows that the airport is located in \textit{``Michigan''} and is near the \textit{``Crooks Road''}. Then we still require the commonsense knowledge that a road (Crooks Road) is a rather small location compared to a state (Michigan). Finally, we can conclude that \textit{``Crooks Road''} locates in \textit{``Michigan''}.

The second case requires the commonsense knowledge about the church. Specifically, if a pope (Benedict XVI) can remove a priest (Maciel) from the ministry, they must be in the same church and hence share the same religion. From sentence [2] we know the priest, Maciel, is a Catholic, hence the pope, Benedict XVI, must also be a Catholic.
Even though our prediction on extracted evidence is correct, the confidence is still not high, leading to the incorrect final prediction.
As the logic chain of commonsense reasoning is always complicated, it is not easy to find a very similar pattern in the training set, or even during pre-training, which makes the problem difficult for a model.

In most of the cases (5 out of 6) in ``\textit{\textbf{Error Type 3}: Fail in Coreferential Reasoning}'', human can still identify the correct relation based on the extracted evidence only.
As shown in our example in Table~\ref{tab:error_case}, in the first sentence, the model wrongly predicts \textit{``Giacomo Casanova''} as the father of \textit{``Manon Balletti''}, but her real father should be an \textit{``Italian actor performing in France''}.
It shows that even the reasoning within a single sentence can be difficult.

Similarly, the example in ``\textit{\textbf{Error Type 4}}'' also shows that the prediction can still be wrong even if we extract the correct evidence sentences and simplify the problem to sentence-level RE.
This suggests that if the performance of sentence-level RE is improved, the performance of DocRE will also improve.

Finally, as described by ``\textit{\textbf{Error Type 5}}'', some examples require direct reasoning from the surface names of the head and tail entities. As shown in the the last case in Table~\ref{tab:error_case}, humans can directly identify that \textit{``China''} is the country of \textit{North China} without reading the document, despite that there are no clue in the document indicates this relation. However, most DocRE models, including \ours, learn to predict the relations only based on the given document and sometimes fail in such cases.

\end{document}